\documentclass[10pt,journal]{IEEEtran}
\usepackage{amsmath}
\usepackage{bm}
\usepackage{graphicx}
\usepackage{float}
\usepackage{amssymb}
\usepackage{amsthm}
\usepackage{authblk}
\usepackage{subcaption}

\begin{document}
\title{Pattern Dependence Detection using $n$-TARP Clustering}

\author{Tarun Yellamraju}
\author{Mireille Boutin}
\affil{Purdue University}
\renewcommand\Authands{ and }

\maketitle

\begin{abstract}
Consider an experiment involving  a potentially small number of subjects. Some random variables are observed on each subject: a high-dimensional one called the ``observed" random variable, and a one-dimensional one called the ``outcome" random variable. We are interested in the dependencies between the observed random variable and the outcome random variable. We propose a method to quantify and validate the dependencies of the outcome random variable on the various patterns contained in the observed random variable. Different degrees of relationship are explored (linear, quadratic, cubic, ...). This work is motivated by the need to analyze educational data, which often involves high-dimensional data representing a small number of students.
Thus our implementation is designed for a small number of subjects; however, it can be easily modified to handle a very large dataset.  
As an illustration, the proposed method is used to study the influence of certain skills on the course grade of students in a signal processing class. A valid dependency of the grade on the different skill patterns is observed in the data.
\end{abstract}

\section{Introduction}
This work is concerned with the problem of finding dependencies between random variables using samples. Many existing methods to address this problem seek to identify a functional relationship between random variables through some type of regression. Linear relationships are by far the easiest to analyze. In the case where the number of random variable samples is significantly larger than the number (dimension) of random variables considered, this can be done by estimating the pairwise correlations between the variables through the covariance matrix. Non-linear relationships can be found by considering non-linear combinations of the random variables, thus increasing the dimensionality of the problem. We are interested in the case where few samples are observed, which can lead to ill-conditioning in the underlying regression, especially when looking for non-linear dependencies. Ill-conditioning can be addressed through the addition of a prior model or assumption for the dependencies, which typically adds a  bias to the solution estimate. For example, Ridge regression \cite{ridge_regression} and LASSO \cite{lasso} do this by promoting sparsity in the functional relationship. 

In this paper, we take a more general standpoint and rephrase the problem of finding functional relationships as 1) finding patterns (clusters) among the samples of an observed random variable, and 2) determining to what extent the outcome random variable depends on the cluster from which the observed random variable is drawn. This point of view allows us to investigate datasets of arbitrarily large dimensions, even small datasets of 20-30 subjects.  

Our work is motivated by the problem of analyzing data arising in educational research. Often, the number of students observed is very small and the data representing each student is high-dimensional. For example, one problem of interest is to determine how student characteristics influence the course outcomes. We view the student characteristics (e.g., attendance rate) as an observed random variable ${\bf x}\in \mathbb{R}^p$, and the course outcomes (e.g., grade) as the outcome variable ${\bf z}\in {\mathbb R}$. The question we seek to answer is ``Do different student characteristics patterns yield different student outcomes?" 

We begin with probing for linear dependencies of the outcome variable on the patterns of observed random variables. This is done by first clustering the feature vectors representing samples of the observed random variable in a large number of different ways. Clustering is accomplished using a proposed modification of the $n$-TARP classifier \cite{TarunBenchmark2016}  and the RP1D clustering method \cite{BoutinCluster2016}. This clustering is based on projecting the sample data onto a random one-dimensional linear subspace and dividing the projected sample data into two by thresholding. Previous work \cite{han2015hidden,BoutinCluster2016} has shown that there exists a significant amount of hidden structure in real datasets that can be uncovered through 1D random projections. Running this random clustering several times yields a large number of binary clusterings. A statistical permutation test is performed to check the validity of each clustering, and non-valid ones are discarded. This clustering approach differs significantly from mainstream ones (e.g., \cite{you2016scalable,you2016oracle} 
) which seek to find a unique ``best'' clustering by numerical optimization. The underlying prior model is also fundamentally different.

The dependencies of the outcome variable is analyzed by looking at the cumulative distribution function (CDF) of the difference between the distribution of the outcome variables between the two subject groups defined by a clustering. The resulting curve is compared to a null hypothesis in order to detect the difference levels at which there is a valid pattern dependence. Higher order dependencies are analyzed by first prolonging the feature vector space to include higher degree monomials before clustering. The influence of individual variables is explored by sequential removal. 
A numerical experiment illustrating the use of our method on data from a course on digital signal processing \cite{Tarun2017Habits,TarunHabitsJournal2017} with a very small number of students is presented at the end of this paper.

\section{Method}
\label{sec:theory}
We begin by describing the first step of our method, which is used to determine the clusterability of the data and thus whether our framework is applicable. Our proposed clustering method is then described, followed by the construction of the CDF curves used to analyze the pattern dependencies and the null hypothesis used for detection. We finish with describing the extension of the feature space to analyze non-linear dependencies as well as the feature analysis process. 

\subsection{Clusterability Quantification}
Our method is applicable when the distribution of the observed random variable has enough structure so that a  projection onto a random line has a high-likelihood of uncovering a binary clustering. To check if this is the case, one should estimate the clusterability of the dataset by plotting the histogram of ``normalized withinss" (${\bf W}$)\cite{BoutinCluster2016}. Normalized withinss is derived from the between-cluster scatter \cite{wang2011ckmeans}. It measures to what extent a 1D set of samples is divided into two clusters. A good binary clustering corresponds to a low value of normalized withinss. Empirically, the probability density function of  ${\bf W}$ should have a non-negligible mass below $W=0.36$ \cite{BoutinCluster2016}.

\subsection{$n$-TARP Clustering}
Our proposed clustering method is non-deterministic and seeks to generate a large number of statistically significant binary clusterings. A fraction of the samples is used to find a clustering (training), and the remaining samples are used to test the validity of the clustering (testing). These two steps are repeated several times until a large number of valid clusterings are obtained. A MATLAB implementation is available \cite{PURR2973}. 


\noindent{\bf Training:}
Let $x_1,\ldots, x_{m_1} \in \mathbb{R}^p$ be the training points.

\begin{enumerate}
\item For $i = 1$ to $n$:
\item Generate a random vector $r_i$ in $\mathbb{R}^p$;
\item Project each $x_j$ onto $r_i$ by taking the dot product $(x_j \cdot r_i)$;
\item Use $2$-means to find 2 clusters in the projected $x_j$;
\item Compute the normalized withinss $W_i$ for this clustering;
\item End loop.
\item Store the vector $r^*$ associated with the smallest $W_i$.
\item Compute and store the threshold $t^*$ separating the two clusters;
\end{enumerate}

\noindent{\bf Validity Testing:}
Let $y_1,\ldots, y_{m_2} \in \mathbb{R}^p$ be the testing points.
\begin{enumerate}
\item Import $r^*$ and $t^*$ from the training phase;
\item Project each testing point $y_j$ onto the vector $r^*$ by taking the dot product $(y_j \cdot r^*)$;
\item Use the threshold $t^*$ to assign a cluster to each of the $y_j$;
\item Perform permutation test with Monte-Carlo simulations \cite{ernst2004permutation} on the projected test data at statistical significance level of 99\%.
\end{enumerate}

There are many ways to generate the random vectors $r_i$. For simplicity, each coordinate is generated using a i.i.d uniform random probability model on $[-1,1]$. There are also many ways to pick an appropriate threshold $t^*$. For simplicity, 
we compute the threshold as the halfway point between the extreme ends (the closer pair) of the projected values from the two classes. Note that, each time the algorithm is run, a different random vector is generated. Therefore, the criteria (feature) used to cluster is different every time. Note also that checking the statistical significance of the grouping using a permutation test is appropriate when the data set is small; for larger datasets, another test should be used.




\subsection{CDF of difference between outcome variable distribution}
Let $\Delta$ be a measure of the difference between two 1D random variable distributions for the outcome variable \textbf{z}. For simplicity, we set $\Delta$ to be the absolute value of the difference between the average values of the two distributions, but other, potentially more accurate measures can be used. We compute  $\Delta$ for each of the previously obtained (valid) clusters.  
Since the clustering process is random, $\Delta$ can be viewed as a random variable.
We estimate its CDF using samples, specifically the values of  $\Delta$ for each of the previously obtained (valid) clusters.



\subsection{Comparison with Null Hypothesis}
The CDF curve is compared to a null-hypothesis CDF curve corresponding to the case with no relationship or dependency between the outcome variable and the clustering of the subjects. This CDF curve is obtained by randomly partitioning the outcome variable values for the entire set of subjects into two groups and computing $\Delta$ for these groups. This partitioning is repeated several times and the resulting sample values of $\Delta$ are used to estimate a CDF curve. The size of the random groups are chosen to match the sizes of the groups obtained by clustering in the previous step. 

The CDF curve obtained as described above will vary from one trial to the next. Thus we compute the average CDF curve, as well as the CDF curves lying one and two standard deviations above/below the average curve. Let us fix a value of $\Delta=\Delta_0$; the values of all the curves obtained through our trials can be used to estimate the exact probability $\alpha_0$ that a curve value at $\Delta_0$ would be below two standard deviations under the mean value. For example, if the distribution of the curve values were a Gaussian, then $\alpha_0$ would be $0.025 ~(2.5 \%)$; 
In general, $\alpha_0$ will tend to be very small.

The values of $\Delta$ at which the experimental CDF curve lies two standard deviations below the average null-hypothesis curve are highlighted; if $\Delta_0$ is a value in the highlighted region, and if the experimental CDF curve value at $\Delta_0$ is $p$, then $p$ of the patterns have a difference of $\Delta_0$ or less, and thus $1-p$ of the patterns have a difference of at least $\Delta_0$. This conclusion is valid with probability $1-\alpha_0$.




\subsection{Feature Space Extension}
The above steps can be applied either directly to the feature vector or, in an identical manner, to an extended feature vector. For example, we can extend the dimensionality of the feature vector $(f_1,f_2,...,f_p)$ by including terms of order $k$ of the form $f_{i_1} \cdot f_{i_2} \cdots f_{i_k} ~\forall~ i_1,i_2,\cdots i_k \in \{1,2,...p\}$, in order to check for non-linear dependencies. 
While extending the feature space in this manner may result in a large increase in the dimensionality of the observed random variable, such high-dimensions are not an issue for our clustering method.




\subsection{Feature Selection}
Another interesting question is whether the outcome variable is dependent on individual features of the observed random variable. This can be investigated by repeating the construction of the CDF curve after removing the feature of interest: if the CDF curve moves up, then the outcome variable dependency on the patterns has been decreased by the removal of the feature, and thus it must be dependent on that feature. If the upward movement is not seen with the degree one curve but with, say, the degree two curve, then the dependency must be quadratic. And so on.


\section{Experimental Results}
We implemented our method in MATLAB and used an educational research 
\cite{Tarun2017Habits} dataset to illustrate the use of our method. The data relates to certain skills (``Habits of Mind'') of students in a course on signal processing. Details of how the data was acquired and a feature vector formed can be found in \cite{TarunHabitsJournal2017}. In this dataset, we have 27 students, each represented by a 26 dimensional feature vector representing skills (observed random variable). The course grade of each student is the outcome variable.





\subsection{Check for Clusterability Results}
\begin{figure}[ht]
\centering
\includegraphics[width=0.5\textwidth]{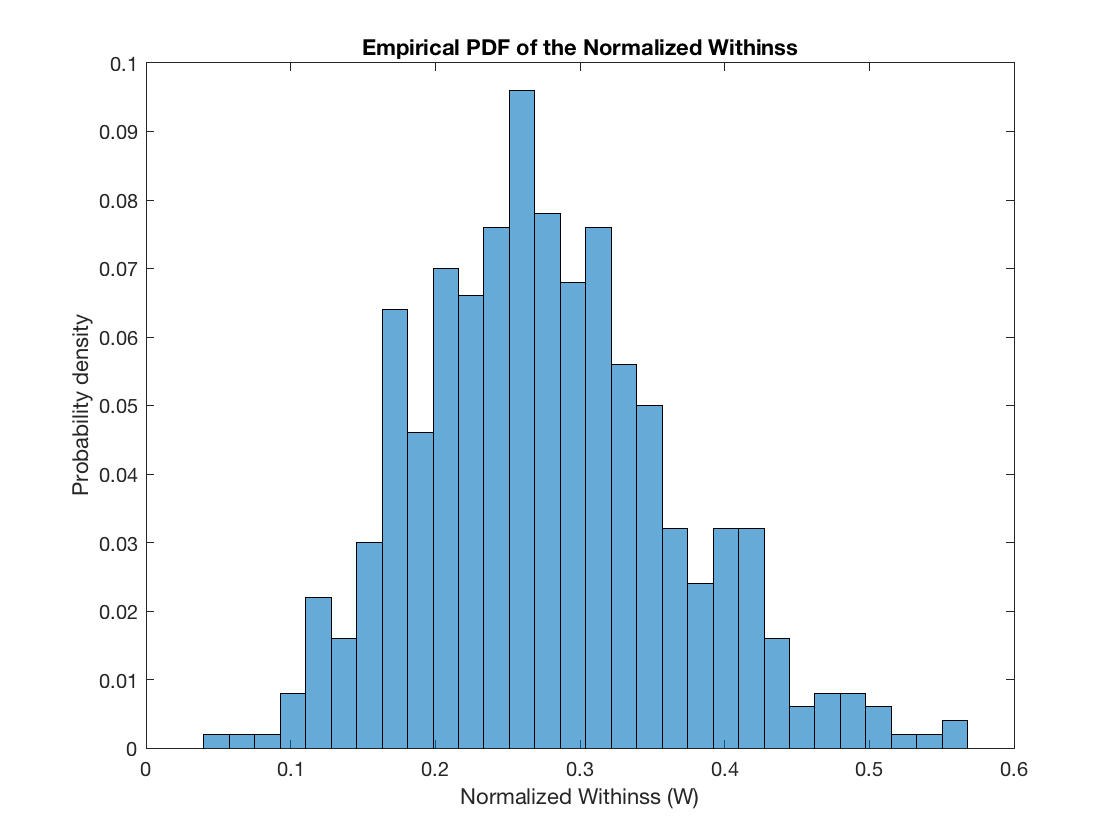}
\caption{Probability density function of normalized Withinss.}
\label{fig:W_pdf} 
\end{figure}

We evaluated the normalized withinss of the cluster assignments resulting from 500 random projection attempts on a random subset of roughly half the data. As seen from Figure \ref{fig:W_pdf}, we observe that nearly 80\% of random projection attempts made on our experimental dataset result in a value of the withinss below 0.36 which is an indicator of the presence of several (linear) binary clustering.

\vspace{-0.5cm}


\begin{table*}[ht]
\centering
\caption{Comparison of nature of clusters formed by different methods using order 1 linear features (26 dimensions)}
\begin{tabular}{|c|c|c|c|c|c|}
\hline
Method & Distinct Clusters & Stat Sig in high D &  Stat Sig in high D & Stat Sig in proj space & Stat Sig in proj space \\
& \% (\#) & (without repetitions) & (with repetitions) & (without repetitions) & (with repetitions) \\
\hline
$k$-means & 2.60 \%  (52) & 38.46 \% & 22.7 \% & N.A. & N.A. \\
$n$-TARP & 74.00 \% (1480) & 16.61 \% & 16.2 \% & 100 \% & 89.6 \% \\
Proclus \cite{proclus} & 100\% (4) & 25\% & 25\% & - & - \\
Clique \cite{clique}& $\phi$ & - & -& - & - \\
Doc \cite{doc} & $\phi$ & - & -& - & - \\
Fires \cite{fires}& $\Phi$ & - & -& - & - \\
INSCY \cite{inscy} & $\phi$ & - & -& - & - \\
Mineclus \cite{mineclus} & $\phi$ & - & -& - & - \\
P3C \cite{p3c} & $\phi$ & - & -& - & - \\
Schism \cite{schism} & $\dagger$ & - & -& - & - \\
Statpc \cite{statpc} & $\dagger$ & - & -& - & - \\
SubClu \cite{subclus} & $\dagger$ & - & -& - & - \\
\hline
\end{tabular}

$\phi$: No clusters were formed, $\Phi$: Overlapping clusters formed, $\dagger$: Did not result in any clusters due to errors over many attempts
\label{table:clus_comp}
\end{table*}

\subsection{$n$-TARP clustering Results}
We set $n=500$ and ran the clustering algorithm a total of 1000 times. For each individual run of the clustering, we randomly split the students into one group of 13 for training, and one group of 14 for validating. The vast majority (74\%) of the clusters found turned out to be distinct (Table \ref{table:clus_comp}). 

In comparison, we also ran $k$-means \cite{kmeans} with $k=2$ a total of 1000 times on the entire dataset. Unsurprisingly, distinct clusters were only obtained 2.6\% of the time out of 1000 attempts (52 different clusterings). We checked for the statistical significance of these clusters (in the original 26 dimensional space) using the high-dimensional version of the permutation test with Monte-Carlo simulations \cite{ernst2004permutation} and found that 38.5\% of these were statistically significant in the original high-dimensional space. So overall, only 20 distinct and statistically significant clusterings were obtained with this method.  



We also compared with several other clustering methods, with a focus on those specially designed for high-dimensions. Our results are summarized in Table \ref{table:clus_comp}. We used the implementations found in Weka \cite{muller2009opensubspace} for these algorithms and consequently could not automate statistical significance testing or make modifications to evaluate clusterability in projected sub-spaces of the algorithms since we did not have access to the source code. Most of the algorithms did not work or produced overlapping clusters as seen in the Table. Proclus was the only method to produce non-overlapping clusters, and in 4 experiments, only 1 cluster assignment was found to be statistically significant in the original high-dimensional space. Such a performance is not very surprising since we are dealing with a very small number of data points (27) in a high-dimensional space (26 D).
As far as we know, our proposed clustering method is the only one that can reliably produce a large number of statistically significant and/or distinct clusters for such a small dataset.

\subsection{CDFs of $\Delta$ and dependency analysis results}

We obtained the empirical CDFs of the absolute difference in average grades of the clusters formed. Specifically, we run our clustering method 10000 times to generate 10000 cluster assignments and retain only those that were statistically significant in the projected one dimensional space. The resulting empirical CDF is shown in Figure 
\ref{fig:nullhyp1} in dotted black. For example, about 30\% of the clusters found by $n$-TARP are associated with an average grade difference of at least 0.5. This is because the y-axis value is about 0.7 at the x-axis value of 0.5. Note that the value of the CDF curve at that point is below the two-sigma line, and thus the value $\Delta=0.5$ is within the significance region. Therefore, there is a statistically valid dependency of the outcome variable at that point (yielding a difference in average mean of at least 0.5),  at the confidence level corresponding to two standard deviations below the mean.

\subsection{Feature Space Extension}
\begin{figure*}[ht]
\begin{subfigure}{.33\textwidth}
  \centering
  \includegraphics[width=.95\linewidth]{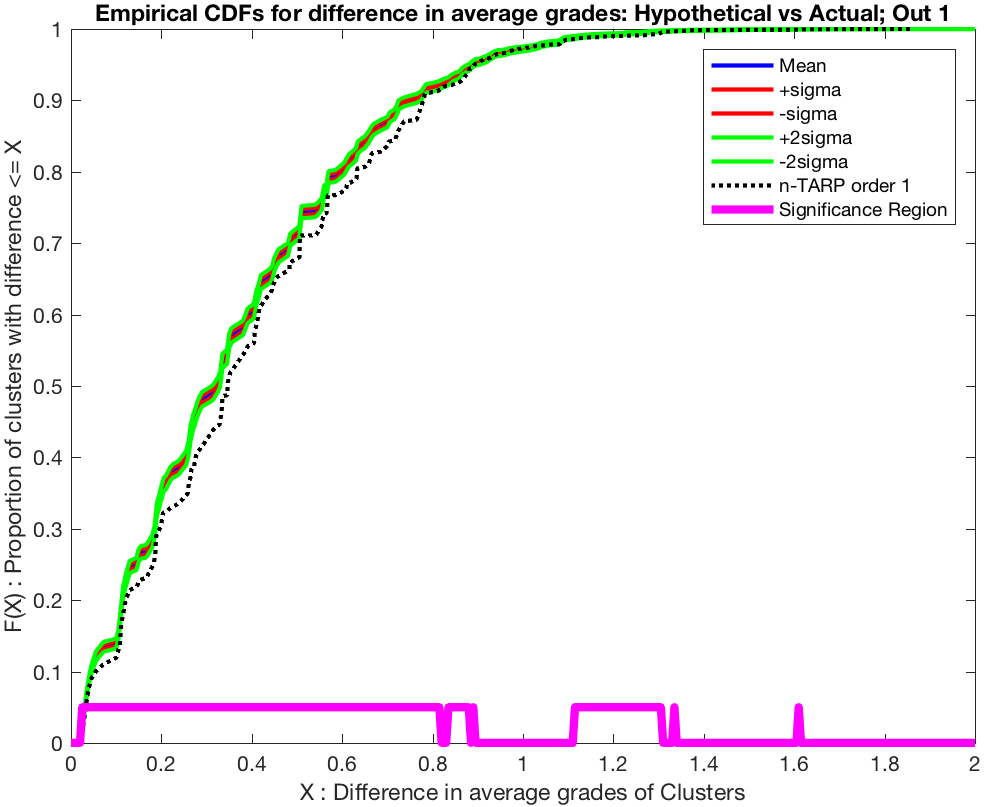}
  \caption{$n$-TARP order 1 vs Null Hypothesis}
  \label{fig:nullhyp1}
\end{subfigure}
\begin{subfigure}{.33\textwidth}
  \centering
  \includegraphics[width=.95\linewidth]{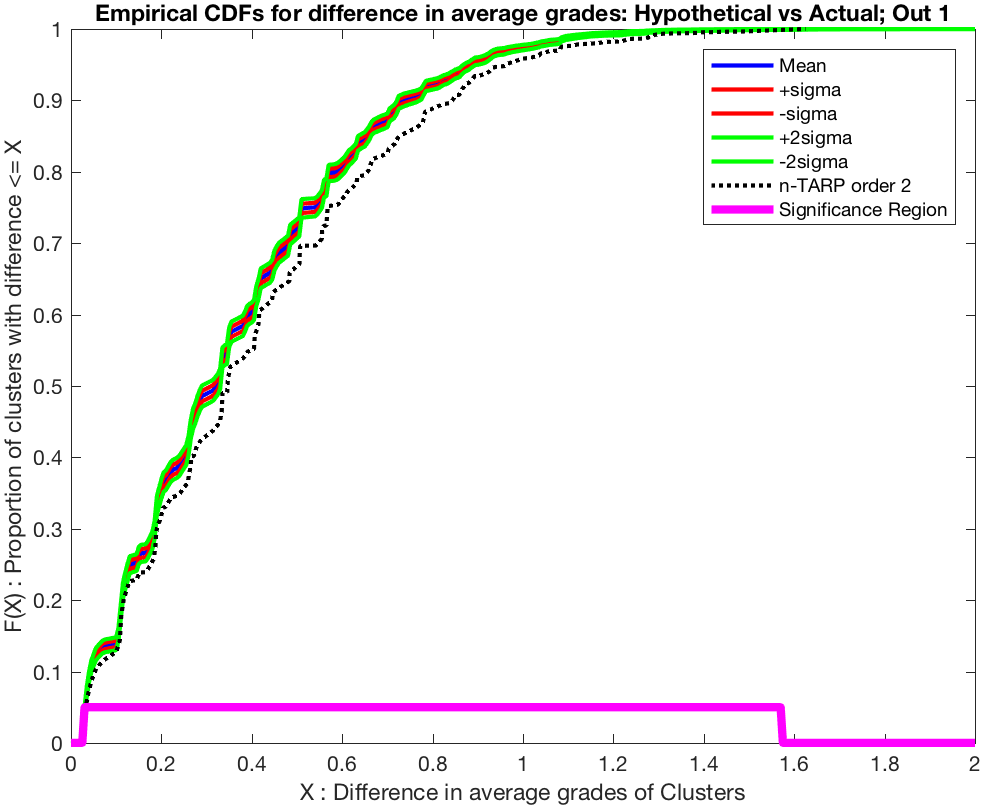}
  \caption{$n$-TARP order 2 vs Null Hypothesis}
  \label{fig:nullhyp2}
\end{subfigure}%
\begin{subfigure}{.33\textwidth}
  \centering
  \includegraphics[width=.95\linewidth]{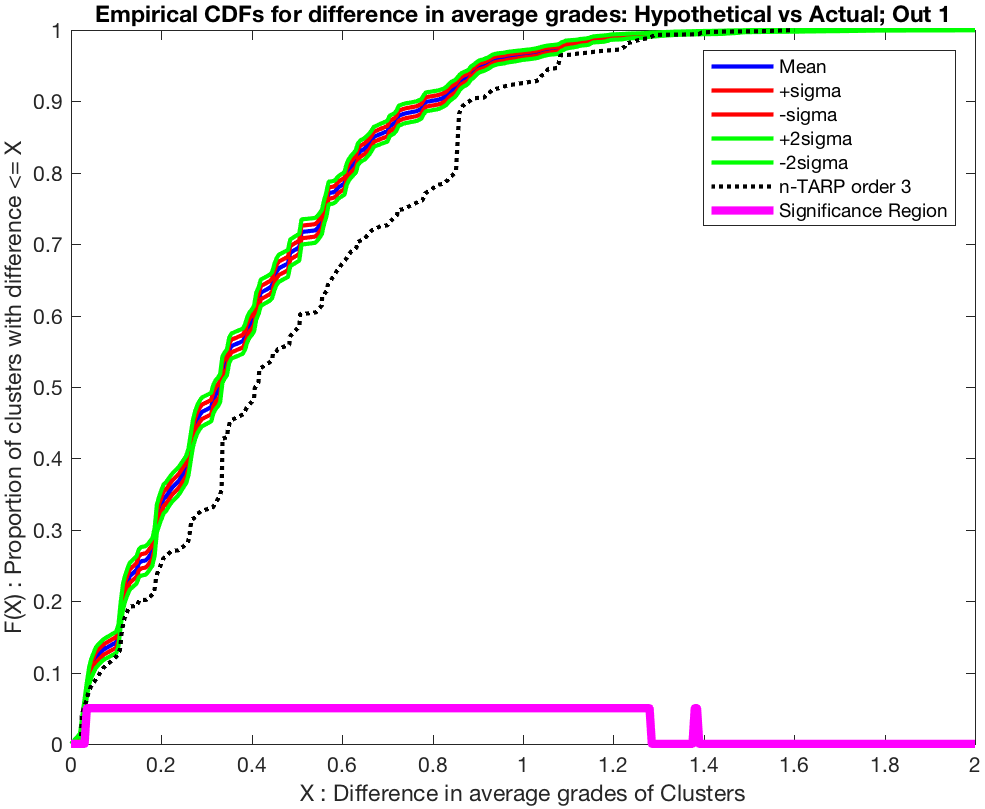}
  \caption{$n$-TARP order 3 vs Null Hypothesis}
  \label{fig:nullhyp3}
\end{subfigure}
\caption{CDF comparison between $n$-TARP and corresponding null hypotheses }
\label{fig:new_nullhyp}
\end{figure*}

We extended our feature vectors by including terms of order two and three, growing the space dimension to 377 and 3003, respectively. The resulting CDF curves are shown in Figures 
\ref{fig:nullhyp2} and \ref{fig:nullhyp3}. Notice how the empirical CDF curve (dotted) moves further down as the degree increases. The significance region, however, is largest when only features of degree two are included.


\begin{figure*}[ht]
\begin{subfigure}{.33\textwidth}
  \centering
  \includegraphics[width=.95\linewidth]{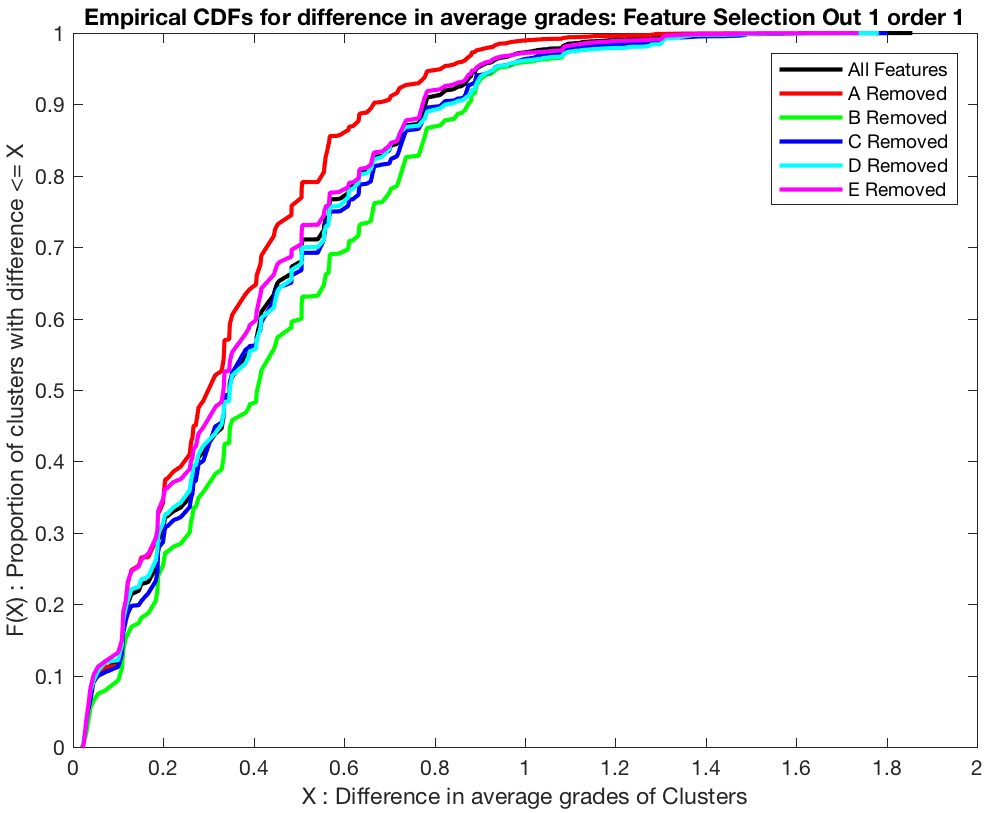}
  \caption{order 1 $n$-TARP}
  \label{fig:FS1}
\end{subfigure}%
\begin{subfigure}{.33\textwidth}
  \centering
  \includegraphics[width=.95\linewidth]{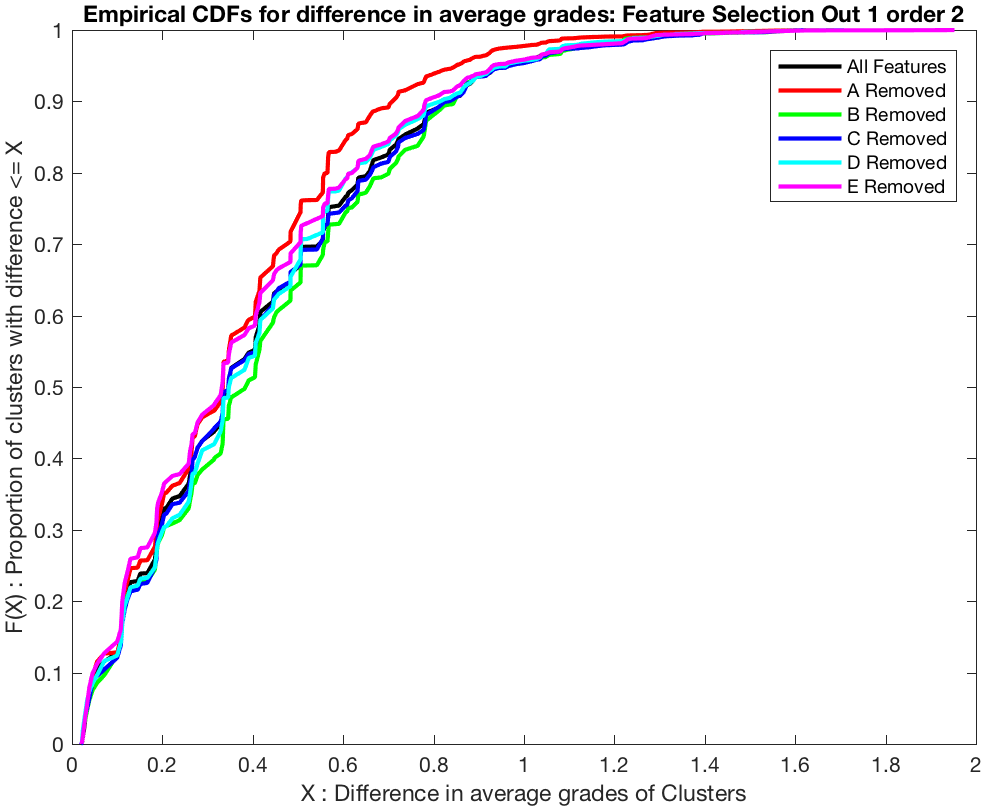}
  \caption{order 2 $n$-TARP}
  \label{fig:FS2}
\end{subfigure}%
\begin{subfigure}{.33\textwidth}
  \centering
  \includegraphics[width=.95\linewidth]{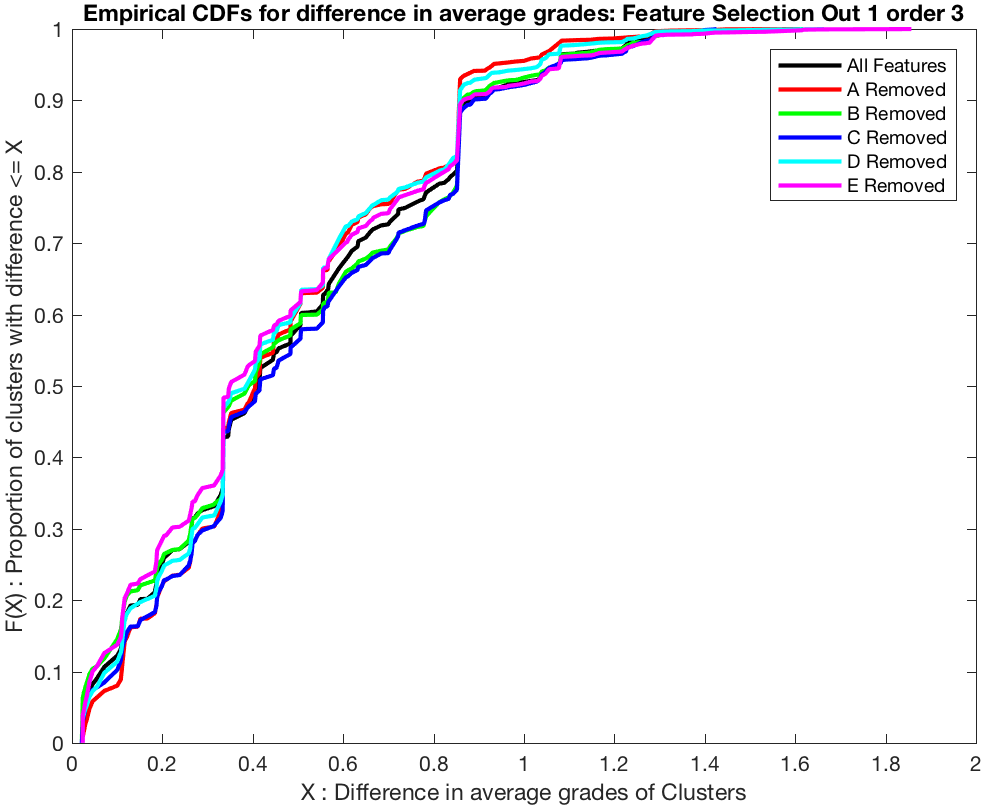}
  \caption{order 3 $n$-TARP}
  \label{fig:FS3}
\end{subfigure}
\caption{Feature Selection CDFs }
\label{fig:featsel}
\vspace{-0.5cm}
\end{figure*}

\subsection{Feature Selection}


Each of the features of our data points is related to one of 5 different skills, which we denote as A,B,C,D and E. We investigate whether the outcome is dependent on a given individual skill by removing the feature coordinates corresponding to that skill and recomputing the corresponding CDF curves. The results using the initial set of features (linear relationship), shown in Figure 
\ref{fig:FS1}, show that the only skills on which the grade depends in a linear fashion are A and, to a lesser extent, E.

The same analysis was carried out after order 2 and order 3 terms expansion of the feature space, respectively (Figures 
\ref{fig:FS2} and \ref{fig:FS3}). For order two, we see that removing A and E seems once again to shift the curve higher. However for order three, removing D also shifts the curve upward. Thus the course grade is dependent on skill D at a degree level three, but not at a degree one (linear) or two.

\section{Conclusions}

We proposed a data analysis method to study the dependence of an outcome random variable on patterns of observed variables. In the method, the existence of dependence patterns is statistically validated and the extent of the dependency is statistically quantified. Changes in the extent of the influence as the feature space is expanded to non-linear terms and as observation variable components are removed are observed. Our approach is computationally inexpensive, scalable to high dimensions, and can be easily modified to handle both very small and large datasets.

As an illustration, we studied 27 subjects whose skills were represented by a 26 dimensional vector in a course on digital signal processing, and analyzed the statistical dependence between these skills and the course grade. 

\textbf{Acknowledgement:} This work was funded in part by NSF grant EEC-1544244.

\bibliographystyle{IEEEtran}
\bibliography{refs_tarun}

\end{document}